\documentclass[letterpaper, 10pt, conference]{ieeeconf} 

\IEEEoverridecommandlockouts
\makeatletter
\newif\ifanonsub
\anonsubfalse
\ifanonsub
  \def\@author{Anonymous Authors}
  \gdef\thanks#1{}
  \def\@thanks{}%
\fi
\makeatother

\usepackage{amsmath}
\usepackage{amssymb}
\usepackage{graphicx}
\usepackage{multirow}
\usepackage{makecell}
\usepackage{array}
\usepackage{dblfloatfix}
\usepackage{tikz}
\usepackage{adjustbox}
\usepackage[T1]{fontenc}
\usepackage[utf8]{inputenc}
 \usepackage{booktabs}
\usepackage{textcomp}
\usepackage{capt-of}
\PassOptionsToPackage{most}{tcolorbox}
\usepackage{listings}
\usepackage{tcolorbox}
\usepackage[table,dvipsnames]{xcolor}
\usepackage{colortbl}
\usepackage{siunitx} 
\usepackage{subcaption}
\usepackage{float}
\usepackage{algorithm}
\usepackage{algpseudocode} 
\algnewcommand\algorithmicinput{\textbf{Input:}}
\algnewcommand\INPUT{\item[\algorithmicinput]}
\algnewcommand\algorithmicoutput{\textbf{Output:}}
\algnewcommand\OUTPUT{\item[\algorithmicoutput]}

\tcbuselibrary{listings,skins,breakable}
\tcbuselibrary{listingsutf8}
\tcbset{listing engine=listings}
\newtcblisting{osmAGlisting}{%
  listing only,
  enhanced,
  colback=black!2,
  colframe=black!20,
  boxrule=0.5pt,
  arc=1pt,
  left=2pt, right=2pt, top=1pt, bottom=1pt,
  boxsep=0pt,
  capture=minipage,
  listing options={%
    basicstyle=\ttfamily\fontsize{6.5pt}{7.5pt}\selectfont\color{osmblue},
    stringstyle=\color{osmgreen},
    showstringspaces=false,
    comment=[l]{//},
    commentstyle=\color{osmgray}\itshape,
    columns=fullflexible,
    keepspaces=true,
    breaklines=true,
    breakatwhitespace=true
  }%
}

\usepackage{float}
\usepackage{placeins}
\title{\LARGE \bf HaltNav: Reactive Visual Halting over Lightweight Topological Priors for Robust Vision-Language Navigation}

\ifanonsub
  \author{Anonymous Authors}
\else
 \author{Zihui Yu$^{1,*}$, Pingcong Li$^{1,*}$, Bichi Zhang$^1$, and Sören Schwertfeger$^{1,\dagger}$%
  \thanks{This work was supported by the National Natural Science Foundation of China under Grant W2531052 and the Key Laboratory of Intelligent Perception and Human-Machine Collaboration (ShanghaiTech University), Ministry of Education, China. The experiments were supported by the Core Facility Platform of Computer Science and Communication, SIST, ShanghaiTech University.}%
  \thanks{$^1$ The authors are with the Key Laboratory of Intelligent Perception and Human-Machine Collaboration, Ministry of Education, China and the School of Information Science and Technology, ShanghaiTech University, Shanghai, China.}%
  \thanks{$^*$ Equal contribution.}
  \thanks{$^\dagger$ Corresponding author.}
}
\fi
\definecolor{osmblue}{HTML}{1A369D}
\definecolor{osmgreen}{HTML}{1E8B42}
\definecolor{osmgray}{HTML}{888888}
\begin{document}
        \maketitle
        \thispagestyle{empty}
        \pagestyle{empty}
        
        \setlength{\itemsep}{1pt}
        \setlength{\parskip}{0pt}
        \setlength{\parsep}{0pt}

\begin{abstract}
Vision-and-Language Navigation (VLN) is shifting from rigid, step-by-step instruction following toward open-vocabulary, goal-oriented autonomy. Achieving this transition without exhaustive routing prompts requires agents to leverage structural priors. While prior work often assumes computationally heavy 2D/3D metric maps, we instead exploit a lightweight, text-based osmAG (OpenStreetMap Area Graph), a floorplan-level topological representation that is easy to obtain and maintain. However, global planning over a prior map alone is brittle in real-world deployments, where local connectivity can change (e.g., closed doors or crowded passages), leading to execution-time failures.
To address this gap, we propose a hierarchical navigation framework HaltNav that couples the robust global planning of osmAG with the local exploration and instruction-grounding capability of VLN. Our approach features an MLLM-based brain module, which is capable of high-level task grounding and obstruction awareness. Conditioned on osmAG, the brain converts the global route into a sequence of localized execution snippets, providing the VLN executor with prior-grounded, goal-centric sub-instructions. Meanwhile, it detects local anomalies via a mechanism we term Reactive Visual Halting (RVH), which interrupts the local control loop, updates osmAG by invalidating the corresponding topology, and triggers replanning to orchestrate a viable detour. To train this halting capability efficiently, we introduce a data synthesis pipeline that leverages generative models to inject realistic obstacles into otherwise navigable scenes, substantially enriching hard negative samples. Extensive experiments demonstrate that our hierarchical framework outperforms several baseline methods without tedious language instructions, and significantly improves robustness for long-horizon vision-language navigation under environmental changes.
\end{abstract}

\section{INTRODUCTION}

Imagine an indoor robotic guide assisting a visually impaired user: ``Please take me to the restroom.'' In practice, such a request is rarely accompanied by an exhaustive route description. Instead, the robot is expected to leverage structural priors (e.g., a building layout) for long-horizon planning, while remaining responsive to local visual contingencies (e.g., a blocked corridor, a closed door or even splashed water on the floor). This goal-oriented setting is challenging for conventional Vision-and-Language Navigation (VLN), since many VLN formulations and models implicitly depend on dense, step-by-step instructions for reliable execution \cite{anderson2018vision}, while those accepting only abstract goals typically depend on costly online exploration \cite{chaplot2020object} and often struggle to stay efficient and robust in large-scale environments. Moreover, recent progress on MLLM or VLA-style embodied agents highlights strong high-level reasoning, yet robustness still hinges on effective closed-loop monitoring and adaptation when the world deviates from the plan \cite{szot2025generalist}.

A natural solution is to provide a prior map and plan globally. However, many existing approaches rely on dense 2D/3D metric or feature maps that are expensive to build and maintain, and may introduce modality misalignment or stale details \cite{gadre2023cows}. In contrast, we advocate a lightweight, floorplan-level topological prior: osmAG \cite{osmag}, a text-structured Area Graph that encodes rooms and passages (e.g., doors) while remaining easy to obtain and token-efficient for language-based reasoning \cite{osmag_llm, xie2024osmagcomprehension}. Given a concise instruction, we perform efficient room-level planning on osmAG to produce a global route without requiring dense trajectory prompts.

The remaining challenge is brittleness under real-world dynamics. Although architectural layout is stable, local connectivity can change. We therefore propose a hierarchical framework that couples osmAG-based global planning with VLN-based local execution. This combination exploits the map prior of osmAG, while intelligently performing local exploration with the VLN model. An LLM-based Graph Grounded Task Dispatcher (GGTD) converts the global route into a sequence of local execution snippets, providing prior-grounded sub-instructions to an off-the-shelf VLN executor for indoor navigation. To robustly handle discrepancies between priors and execution-time conditions, we introduce a VLM-based Reactive Visual Halting (RVH) mechanism. It monitors egocentric observations during local execution, detects blocked connectivity, interrupts the control loop, updates osmAG by invalidating the corresponding connector, and triggers global replanning to orchestrate a viable detour. This design aligns with emerging views that treat monitoring and failure detection as first-class components for reliable embodied autonomy \cite{zhou2025codeasmonitor}.

Such intelligent scene understanding requires specified VLM reasoning ability. However, training this capability solely from simulated obstacles is limited by the finite diversity of 3D assets and the resulting sim-to-real gap. We thus build an automated failure-injection synthesis pipeline that edits otherwise navigable scenes into obstructed ones using generative models, producing scalable hard negatives for finetuning. Finally, existing VLN benchmarks typically assume no prior map and rely on dense step-by-step instructions, making them ill-suited for evaluating the conflict between a static topological prior and a dynamic environment. To fill this gap, we curate tasks across instruction granularity levels ranging from step-by-step guidance to goal-only commands, and additionally test obstacle-injected settings that deliberately violate the static prior. Our experiments show that integrating lightweight structural priors with VLN-based local execution and RVH mechanism substantially improves long-horizon robustness under environmental changes.

We summarize our contributions as follows:
\begin{itemize}
    \item \textbf{Hierarchical Planning and Execution:} We propose a hierarchical navigation framework named HaltNav that uses lightweight osmAG priors for room-level global planning and leverages off-the-shelf VLN models for local execution.
    \item \textbf{LLM Task Dispatching:} We design an LLM-based task dispatcher that converts a global osmAG route into localized door-to-door execution snippets, enabling concise, goal-centric instructions without dense trajectory prompts.
    \item \textbf{Reactive Visual Halting:} We introduce a finetuned VLM-based obstruction handler that detects blocked connectivity online, updates osmAG by invalidating passages, and triggers global replanning to detour safely.
    \item \textbf{Failure-Injection Data Synthesis:} We develop an automated data synthesis pipeline that generates scalable hard negatives by editing traversable scenes into obstructed ones, improving the obstruction handler’s robustness.
    \item \textbf{Robustness-Centric Benchmark:} Unlike existing VLN benchmarks that assume unknown environments with dense instructions, we construct evaluation tasks where agents access a static osmAG prior but face dynamically injected obstacles. We manually curate high-quality annotations across instruction granularity levels and validate on both simulation and physical robots.
\end{itemize}

\section{RELATED WORK}

\subsection{Vision-and-Language Navigation (VLN)}
Traditional Vision-and-Language Navigation (VLN) requires agents to navigate unseen environments by strictly following fine-grained, step-by-step instructions. While significant progress has been made on benchmarks like R2R and VLN-CE \cite{anderson2018vision, vln_ce}, these detailed instructions are often impractical for real-world deployment, where human users typically provide abstract, goal-oriented commands. Recently, the focus has shifted toward high-level, goal-oriented instruction following. For instance, DV-VLN \cite{li2026dv} proposed a generate-then-verify paradigm to improve the reliability of LLM-based navigation. Furthermore, MapNav \cite{zhang2025mapnav} effectively mitigated the spatio-temporal overhead of continuous navigation by replacing historical frames with Annotated Semantic Maps (ASM). Similar attempts appear in \cite{wu2025map} and \cite{wang2026matchnav}. However, while those methods demonstrated strong performance, constructing and continuously updating visual 2D/3D semantic maps incurs prohibitive computational demands on resource-constrained mobile robots. In contrast, our framework targets abstract instructions but circumvents dense visual reconstruction by leveraging highly lightweight, text-based priors, ensuring scalability in real-world scenarios.


\subsection{Topological Maps for Navigation and LLM Reasoning}
Abstracting environments into topological or scene-graph structures alleviates the computational burden of dense metric mapping and has long supported navigation. Such graphs serve as spatial memory for object-goal and image-goal tasks \cite{liu2025toponavtopologicalgraphskey, yoo2024ovgnav, kim2023topological}, and can be further enriched with geometric cues to support vision-language navigation and traversability-aware control \cite{Liu2023BirdsEyeViewSG, podgorski2025tango}. These representations also align naturally with large language models, which have been used to drive navigation chatbots, track object relations online, and plan over text-serialized scene graphs \cite{song2025guidellme, rana2023sayplan}. In a similar spirit, recent text-based maps \cite{osmag_llm, xie2024osmagcomprehension} show that hierarchical and human-readable representations provide a token-efficient prior for global routing without complex visual-textual alignment. However, a purely text-driven planner operating on static priors intrinsically conflicts with the dynamic nature of physical environments. Our work builds upon the efficiency of osmAG but explicitly addresses its static vulnerability by introducing a dynamic re-grounding mechanism to survive unexpected environmental changes.

\subsection{Hierarchical Control and Reactive Replanning}
The pursuit of long-horizon autonomy has driven a paradigm shift from monolithic policies to hierarchical Vision-Language-Action (VLA) architectures. Frameworks such as Hi-robot \cite{shi2025hirobot} and VLA-OS \cite{gao2025vla} explicitly decouple high-level semantic planning from low-level embodiment grounding, employing an MLLM as a global orchestrator and a separate policy for local kinematics. While this decoupling is effective, executing macroscopic plans in dynamic environments requires robust error-recovery mechanisms. Traditional stacks often suffer from errors when facing impassable obstacles \cite{huang2025unemo}. To address this, ReCAPA \cite{fan2025diffusion} introduces a hierarchical predictive correction framework that anticipates deviations and adjusts representations across actions, subgoals, and trajectories to mitigate cascading failures in long-horizon reasoning. Recent works like EPoG \cite{yang2026integrated} have introduced situated replanning to update belief graphs dynamically. Advancing this philosophy, we introduce a Reactive Visual Halting mechanism. Rather than relying on passive replanning, our local RVH executor performs continuous visual grounding to actively monitor spatial affordances. Upon detecting critical anomalies, it decisively intercepts the local control loop, prompting the MLLM brain to update the osmAG edges and orchestrate a feasible topological detour.

\begin{figure*}[t]
                \centering
                \includegraphics[width=0.85\linewidth]{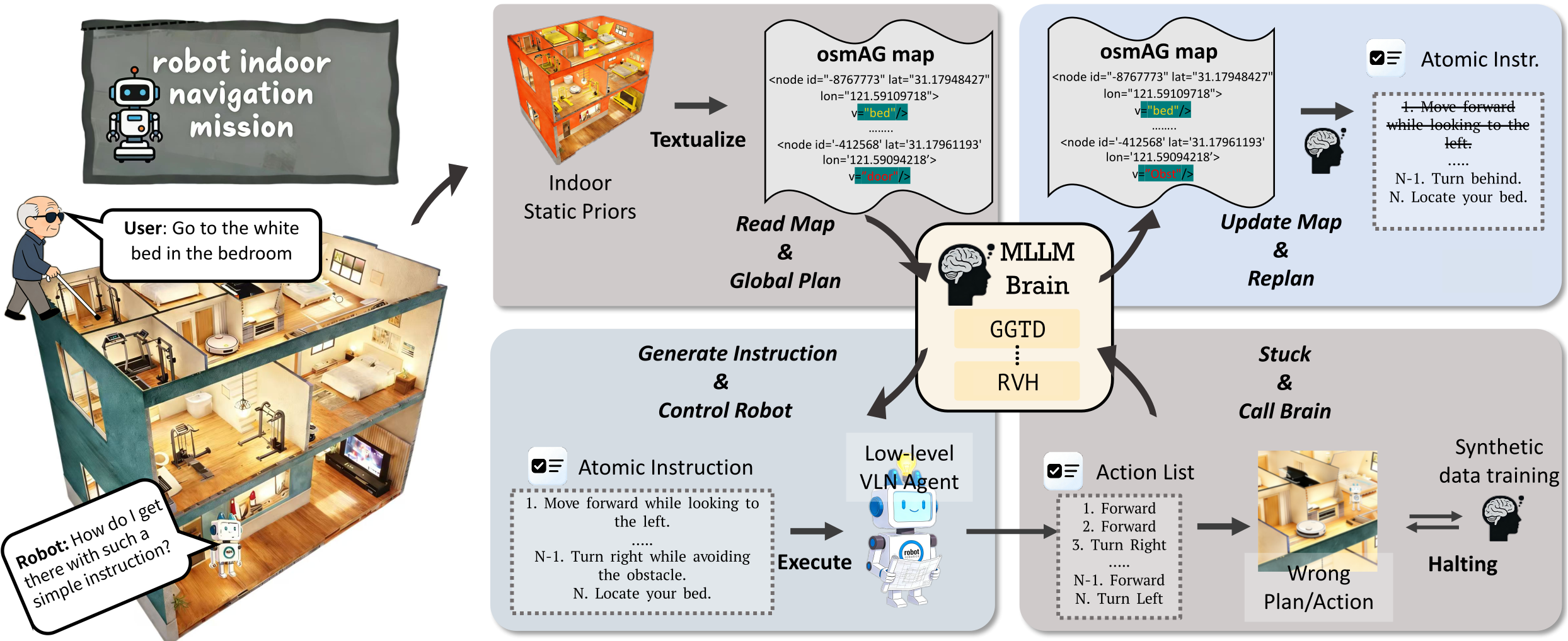}
                \caption{\textbf{Overview of the proposed hierarchical navigation framework.} To handle abstract user commands, our system decouples navigation into global semantic planning and local physical execution. \textbf{(Top Left)} The GGTD (Graph-Grounded Task Dispatcher) in the MLLM brain reasons the text-based static prior (osmAG) to formulate an initial route. \textbf{(Bottom Left)} The overarching plan is broken down into atomic instructions for the low-level VLN agent to execute. \textbf{(Bottom Right)} Upon encountering unmapped dynamic anomalies, the RVH (Reactive Visual Halting) mechanism interrupts the execution loop. \textbf{(Top Right)} The brain subsequently updates the internal map to prune the blocked passage and autonomously replans a collision-free detour.}
                \label{fig:pipeline}
\end{figure*}

\section{METHODOLOGY}
Our proposed hierarchical navigation framework aims to achieve robust, goal-oriented navigation in dynamic environments. As depicted in Fig.~\ref{fig:pipeline}, the core of the system is the MLLM (Multi-modal Large Language Model) brain, decomposing the global path into atomic instructions based on grounded osmAG information. During local execution, the RVH mechanism monitors the input image. Once blockages are spotted, it reactively cuts into the control loop and initiates replanning. The full algorithm is displayed in Algorithm~\ref{algo:Nav}.

\subsection{osmAG Map}
The osmAG agent is responsible for map maintenance and global planning using OpenStreetMap Area Graph (osmAG) \cite{ag} \cite{osmag}, a novel map format that extends the OpenStreetMap (OSM) standard to store hierarchical, semantic and topological information, to represent large-scale environments compactly and semantically. In the map, it represents environments using areas (closed polygons) as nodes and passages (lines shared by two areas) as edges, supporting multi-floor indoor and outdoor scenarios. As the osmAG map can be automatically generated from floorplan or CAD files \cite{osmAG_cad}, acquiring the map prior in the real world is simple, without tedious mapping or reconstruction procedures. 

\begin{figure}[t]
  \centering
  \includegraphics[width=0.45\textwidth]{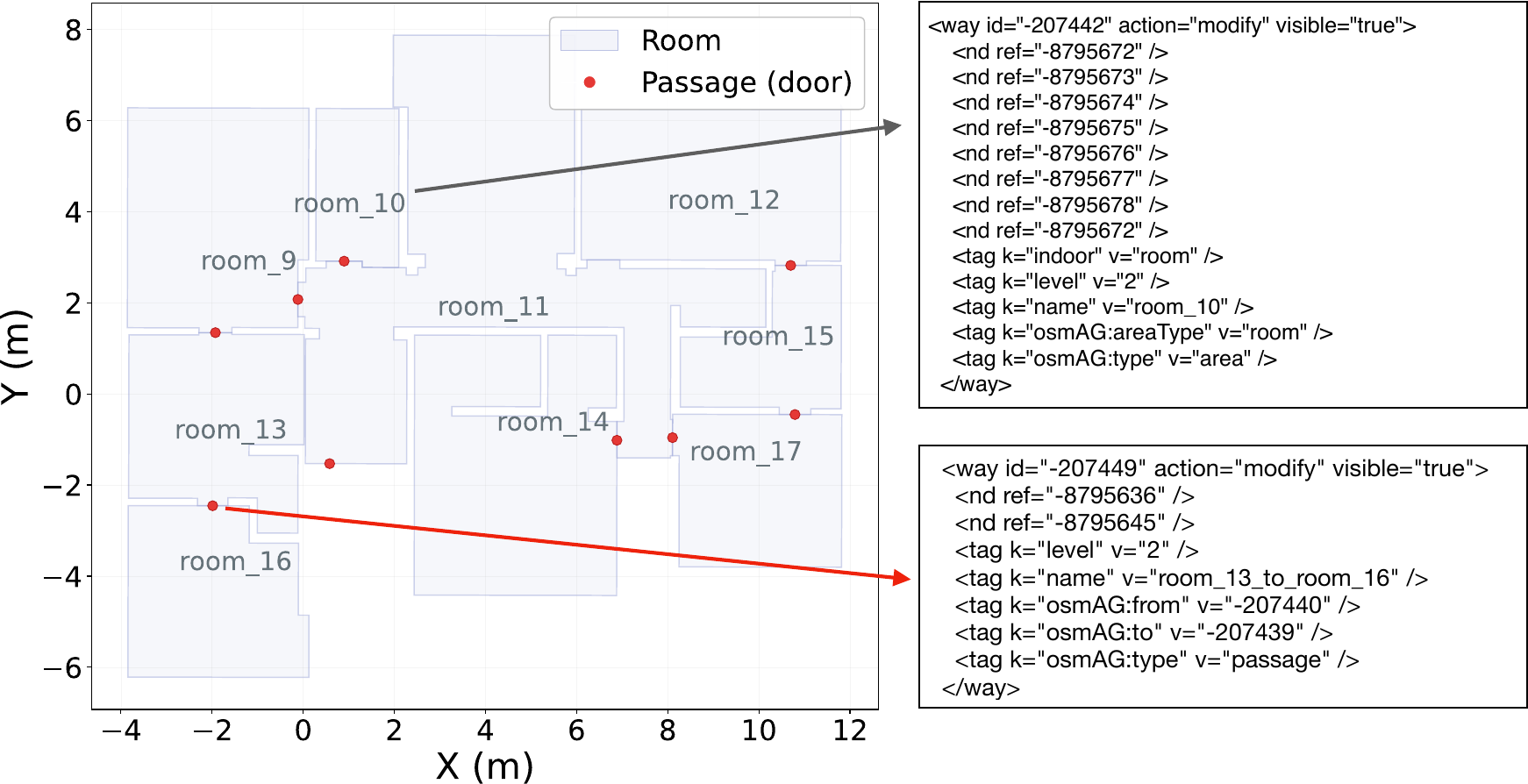}
  \caption{An example of the osmAG map of the second floor of scene 00862 in HM3D dataset \cite{ramakrishnan2021hm3d}. The map only contains the room info and the passage (e.g., doorway), shown in red dots, connecting the rooms.}
  \label{fig:osmag}
\end{figure}

We adopt the global planning framework proposed in the osmAG \cite{osmag}, which constructs a passage-level graph from the area graph and computes traversal costs between passages using A* on rendered 2D occupancy grid sub-maps from the area polygon. These precomputed costs are stored and reused for all queries. This passage-level graph is the actual map maintained by the osmAG agent during navigation. An example of the osmAG map is shown in Fig.~\ref{fig:osmag}.

The robot may encounter unexpected obstacles or changes which are not recorded in the map (e.g., a locked door). The osmAG agent supports online map updates and re‑planning based on visual observations. If a passage is detected as blocked, we remove the corresponding node and its incident edges from the passage‑level graph. If an area becomes non‑traversable (e.g., due to construction), we set the traversal costs of all edges passing through that area to infinity. The planner then automatically avoids these regions in subsequent queries.

The high-level organizer can request re‑planning when the VLN agent reports navigation failures. The osmAG agent recomputes the path using the updated map, providing a new global plan that adapts to the dynamic environment.

\subsection{Problem Formulation}
The long-horizon VLN task can be formulated within the framework of a Hierarchical Semi-Markov Decision Process. The prior is represented by a purely text-based osmAG, defined as a weighted directed dynamic graph $\mathcal{G}_t = (\mathcal{V}, \mathcal{E}, W_t)$. Here, $\mathcal{V}$ denotes the set of semantic nodes (e.g., regions), $\mathcal{E} \subseteq \mathcal{V} \times \mathcal{V}$ represents the traversable passages, and $W_t: \mathcal{E} \to \mathbb{R}^+$ defines the time-varying traversal cost.

Given a high-level instruction $I_{target}$ and an initial observation $o_0$, we decouple the navigation task into a bi-level hierarchy. 
At the macro-level, the global policy $\pi_{high}$, driven by GGTD, operates at discrete steps $i$. It breaks down the overarching goal into a sequence of textual macro-actions $m_i \in \mathcal{M}$, which serve as atomic natural language instructions. 

At the micro-level, the local policy $\pi_{low}$ from an end-to-end VLN network with weights $\theta$ acts as the physical executor at control steps $t$. It directly maps the current egocentric observation $o_t$ and the active macro-action $m_i$ to a discrete motor command $a_t \in \mathcal{A}$ (e.g., move forward, turn left, turn right):
\begin{equation}
    a_t \sim \pi_{\mathrm{low}}\!\left(\cdot \mid o_t,\, m_i;\theta\right),
    \qquad t \in \left[t_i,\, t_i+\Delta t_i\right)
    \label{eq:pi_low}
\end{equation}
where $\Delta t_i$ is the variable execution duration of $m_i$. Crucially, the execution of each macro-action is dynamically governed by a termination indicator $\beta(o_t, m_i)\in\{0,1\}$, which actively determines when to halt the local policy and request the next macro-action.

\subsection{Graph-Grounded Task Dispatcher via osmAG Map}
To avoid complex multi-modal projection overheads, we directly leverage the inherent text-based nature of the osmAG. At macro-step $i$, the local sub-graph surrounding the agent is represented as a structured textual prompt $\mathcal{P}(\mathcal{G}_t)$. The GGTD, which is essentially an LLM model, acts as a task dispatcher, directly reading this text-based map to determine the optimal macroscopic waypoint $m_i$:
\begin{equation}
    m_i = \mathrm{GGTD}(\mathcal{P}(\mathcal{G}_t),\, I_{\mathrm{target}},\, \mathcal{H}_{i-1}),
    \label{eq:ggtd}
\end{equation}
where $I_{\mathrm{target}}$ is the target instruction and $\mathcal{H}_{i-1}$ is the trajectory history. This direct textual prompting naturally aligns the spatial graph routing with the MLLM's inherent reasoning capabilities.

\subsection{Reactive Visual Halting Mechanism}
At the execution level, we deploy a continuous vision-language navigation policy as the local executor $\pi_{\mathrm{low}}$. Its primary role is to ground the macroscopic textual waypoint $m_i$, generated by GGTD, into physical motor commands. To achieve this, the executor processes both the real-time egocentric visual observation $o_t$ and the instruction $m_i$. A cross-modal fusion module aligns the visual spatial features with the linguistic semantics to directly output the next micro-action $a_t$. 

Crucially, $\pi_{\mathrm{low}}$ operates entirely agnostic to the global osmAG $\mathcal{G}_t$. This explicit decoupling relieves the local policy from the computational burden of global spatial memory, allowing it to focus exclusively on local obstacle avoidance and short-term instruction grounding.

Instead of allowing $\pi_{\mathrm{low}}$ to execute $m_i$ in an open-loop manner, we formalize the termination function $\beta(o_t, m_i)$ via the Reactive Visual Halting Mechanism that unifies safety-critical physical signals and semantic infeasibility cues:
\begin{equation}
    \beta(o_t,m_i)=\!\left[
    \underbrace{\sum_{j=0}^{k-1} c_{t-j}}_{\text{collision accumulation}} \ge \tau_c
    \;\;\vee\;\;
    \underbrace{s_{\mathrm{MLLM}}(o_t,m_i)}_{\text{traversability}} 
    \right]
    \label{eq:rvh}
\end{equation}
The first term denotes a \textit{collision-based halting}, where
$c_t\in\{0,1\}$ is a binary collision indicator at step $t$.
The threshold $\tau_c$ bounds the maximum tolerated cumulative collision
frequency within a sliding window of size $k$, acting as a physical
safety net. The second term denotes an \textit{MLLM-based traversability
halting}, where $s_{\mathrm{MLLM}}(o_t,m_i)\in[
\textit{True}, \textit{False}
]$ represents the MLLM-scored traversability, for example identifying
unmapped crowded corridors or closed doors. Once either condition is
met, the system immediately intercepts the local controller and triggers
global topological replanning.

\subsection{Dynamic Topology Belief Updating}
When a reactive halt is triggered, simply appending textual descriptions of the obstacle to the RVH module often leads to context overflow and spatial hallucinations. Instead, we directly intervene in the osmAG map maintained by the osmAG agent. 

Unlike standard node-to-node graphs, the osmAG global planner operates on a passage-level graph, where edges represent the precomputed A* traversal costs between passages (e.g., doors or room boundaries). Let this passage-level graph be denoted as $\mathcal{G}_p = (\mathcal{P}, \mathcal{E}_p)$, which is derived from the original graph $\mathcal{G}_t$, where $\mathcal{P}$ represents the set of passages, and $C_t(p_i, p_j)$ is the stored traversal cost between passage $p_i$ and $p_j$ at time $t$.

When the local executor visually confirms an unmapped, impassable anomaly blocking the route between $p_i$ and $p_j$, we dynamically update the topological belief by penalizing the traversal cost:
\begin{equation}
    C_{t+1}(p_i, p_j) = 
    \begin{cases}
      \infty, & \text{if visual anomaly is detected},\\
      C_t(p_i, p_j), & \text{otherwise}.
    \end{cases}
    \label{eq:update}
\end{equation}
By setting the cost to infinity, the blocked intra-area connection is effectively pruned from the passage-level graph. The global planner then seamlessly recalculates the optimal route over the updated $\mathcal{G}_p$. This explicit modification guarantees a physically viable detour, forcing the agent to bypass the obstruction without relying on unstable prompt engineering.

\begin{algorithm}[t]
\caption{The full pipeline of HaltNav.}
\label{alg:framework}
\small
\begin{algorithmic}[1]
\INPUT
Hierarchical osmAG $\mathcal{G}_t$ with cost matrix $C_t$;
target instruction $I_{\mathrm{target}}$; observations $o_t$;
low-level VLN policy $\pi_{\mathrm{low}}$;
collision indicator $c_t$; window size $k$; max macro steps $I_{\max}$.
\OUTPUT Executed action trajectory $\{a_t\}$.

\State $t \gets 0$, $i \gets 0$, $\mathcal{H}_{0}\gets \emptyset$, collision buffer $\mathcal{C}\gets [\,]$
\State Compute initial global route on $\mathcal{G}_p$ (A* using $C_t$)

\While{$i < I_{\max}$}
    \State Build local textual prompt $\mathcal{P}(\mathcal{G}_t)$ around current agent location
    \State $m_i \gets \mathrm{GGTD}\!\left(\mathcal{P}(\mathcal{G}_t),\, I_{\mathrm{target}},\, \mathcal{H}_{i}\right)$ \Comment{Eq.~\eqref{eq:ggtd}}
    \State \texttt{halt} $\gets$ \textbf{False}
    \While{\textbf{not} \texttt{halt}}
        \State Sample and execute $a_t \sim \pi_{\mathrm{low}}$ \Comment{Eq.~\eqref{eq:pi_low}}
        \State Observe $o_{t+1}$ and collision flag $c_t$
        \State Update collision buffer $\mathcal{C}$ with $c_t$ (keep last $k$ steps)
        \State Evaluate RVH termination $\beta(o_{t+1},m_i)$ \Comment{Eq.~\eqref{eq:rvh}}
        \If{$\beta(o_{t+1},m_i)=1$}
            \State \texttt{halt} $\gets$ \textbf{True}
            \If{a passage $p_u$ is blocked}
                \State Remove $p_u$ and its incident edges from $\mathcal{G}_p$
            \ElsIf{an area becomes non-traversable}
                \State Set costs of corresponding edges to $\infty$ \Comment{Eq.~\eqref{eq:update}}
            \EndIf
            \State Recompute global route on updated $\mathcal{G}_p$ (A*)
        \EndIf
        \State $t \gets t+1$
    \EndWhile
    \State Update history $\mathcal{H}_{i+1} \gets \mathcal{H}_{i} \cup \{(m_i, o_t)\}$; $i \gets i+1$
    \If{\textbf{GoalReached}$(o_t, I_{\mathrm{target}})$}
        \State \textbf{break}
    \EndIf
\EndWhile
\State \Return $\{a_t\}$
\end{algorithmic}
\label{algo:Nav}
\end{algorithm}

\subsection{Data Synthesis for Supervised Finetuning}


\begin{figure}[t]
  \centering
   \includegraphics[width=0.85\linewidth]{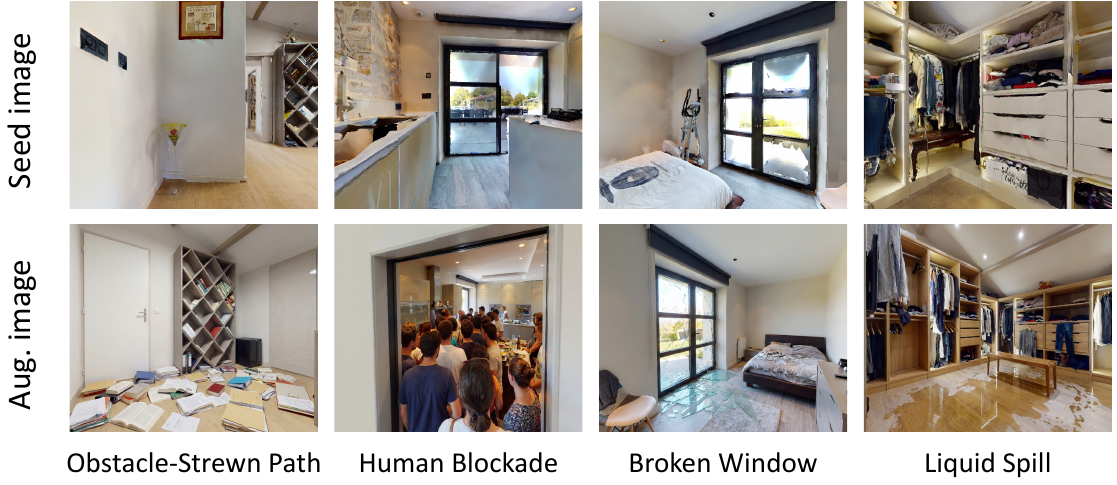}
  \caption{\textbf{Qualitative examples} of Failure-Injected Counterfactual Synthesis. To bridge the sim-to-real gap and generate infinitely diverse anomalies, we leverage generative visual inpainting models to modify RGB observations. The top row shows original Seed images denoting traversable paths. The bottom row displays the resulting Augmented images, where semantic obstacles, such as clutter, crowds, or physical hazards—are realistically synthesized. These serve as high-fidelity negative samples for training the reactive halting capability.}
  \label{fig:data_generation}
\end{figure}

While modern MLLMs inherently possess strong zero-shot visual reasoning capabilities, deploying them off-the-shelf for RVH often yields brittle performance. They typically struggle to ground general visual semantics into robot-specific spatial affordances. For instance, a pre-trained model might recognize a ``cleaning cart," but fail to explicitly judge whether it physically obstructs the current navigation waypoint. Consequently, targeted fine-tuning is essential to align the model for robust anomaly detection. 

However, curating such training data presents a critical bottleneck. Relying exclusively on simulator-based failure injection is inherently constrained by the finite diversity of 3D assets and the inevitable sim-to-real domain gap. To systematically address this limitation and explicitly teach the agent when to halt, we propose a synthesis pipeline that integrates physical simulation with scalable generative data augmentation.

First, the \textit{Physics-Based Engine} operates within the Habitat simulator, where we extract expert navigation trajectories and stochastically inject dynamic 3D obstacles at critical topological bottlenecks. Second, to complement the limited simulated assets, the \textit{Generative Perturbation Engine} leverages pre-trained diffusion models to augment real-world navigation data. Specifically, we apply targeted visual inpainting to originally traversable regions within RGB observations, synthesizing high-fidelity counterfactual anomalies (e.g., unmapped physical blockades or pedestrian crowds). 

As shown in Fig.~\ref{fig:data_generation}, this approach curates a comprehensive instruction-tuning dataset comprising pairs $\mathcal{D} = \{ (X_p, Y_{\mathrm{no\text{-}halt}}), (X_a, Y_{\mathrm{halt}}) \}$. Here, $X_p$ and $X_a$ denote the passable and anomalous observations, while $Y_{\mathrm{no\text{-}halt}}$ and $Y_{\mathrm{halt}}$ are their corresponding target decisions of continuing execution or triggering a reactive halt, respectively. We subsequently employ Supervised Fine-Tuning (SFT) utilizing Low-Rank Adaptation (LoRA) to optimize the MLLM by minimizing the negative log-likelihood:
\begin{equation}
    \mathcal{L}_{\mathrm{SFT}} = - \sum_{(X, Y) \in \mathcal{D}} \sum_{j=1}^{|Y|} \log p_{\mathrm{MLLM}}\!\left(y_j \mid y_{<j},\, X;\, \Theta_{\mathrm{MLLM}}\right)
\end{equation}
Consequently, the MLLM brain acquires the active reasoning capabilities necessary for reactive halting and out-of-distribution (OOD) generalization in complex physical environments.

\begin{table*}[t]
  \caption{Simulation results under standard (B) and obstacle-injected (O) settings across instruction levels (L0--L2). For SR/SPL/OS, Drop$=$B$-$O (lower is better). For NE, $\Delta$=O$-$B (lower is better). \textbf{Bold} marks the best SR, SPL and OS under B and O per level.}
  \label{tab:replanning_bo}
  \centering
  \setlength{\tabcolsep}{4.5pt}
  \renewcommand{\arraystretch}{1.05}
  \begin{tabular}{l|c|ccc|ccc|ccc|ccc}
    \hline
    \multirow{2}{*}{Method} & \multirow{2}{*}{Level} & \multicolumn{3}{c|}{SR (\%)} & \multicolumn{3}{c|}{SPL} & \multicolumn{3}{c|}{OS (\%)} & \multicolumn{3}{c}{NE (m)} \\
    \cline{3-14}
     &  & B & O & Drop & B & O & Drop & B & O & Drop & B & O & $\Delta$ \\
    \hline
    Navid~\cite{zhang2024navid} & L0 & 73.13 & 6.25 & 66.88 & 0.68 & 0.05 & 0.63 & 83.13 & 6.25 & 76.88 & 2.82 & 9.64 & 6.82 \\
     & L1 & 58.13 & 6.25 & 51.88 & 0.52 & 0.02 & 0.50 & 61.25 & 6.25 & 55.00 & 4.24 & 7.87 & 3.62 \\
     & L2 & 49.38 & 0.00 & 49.38 & 0.43 & 0.00 & 0.43 & 56.25 & 0.00 & 56.25 & 5.16 & 8.45 & 3.29 \\
    \hline
    OmniNav~\cite{xue2026omninavunifiedframeworkprospective} & L0 & \textbf{90.63} & 12.50 & 78.13 & \textbf{0.82} & 0.06 & 0.76 & \textbf{96.88} & 18.75 & 78.13 & 1.30 & 9.54 & 8.24 \\
     & L1 & \textbf{80.63} & 18.75 & 61.88 & \textbf{0.76} & 0.13 & 0.63 & \textbf{87.50} & 18.75 & 68.75 & 2.35 & 9.84 & 7.49 \\
     & L2 & 54.38 & 6.25 & 48.13 & 0.47 & 0.04 & 0.43 & 60.00 & 6.25 & 53.75 & 5.02 & 8.96 & 3.94 \\
    \hline
    StreamVLN~\cite{wei2025streamvln} & L0 & 72.50 & 37.50 & 35.00 & 0.62 & 0.19 & 0.43 & 83.75 & 37.50 & 46.25 & 3.06 & 6.61 & 3.55 \\
     & L1 & 63.75 & 18.75 & 45.00 & 0.57 & 0.09 & 0.48 & 68.13 & 25.00 & 43.13 & 4.43 & 9.71 & 5.28 \\
     & L2 & 36.25 & 18.75 & 17.50 & 0.28 & 0.13 & 0.15 & 43.13 & 31.25 & 5.63 & 6.66 & 7.62 & 0.96 \\
    \hline
    Uni-navid~\cite{zhang2024uninavid} & L0 & 75.63 & 6.25 & 69.38 & 0.68 & 0.02 & 0.66 & 81.88 & 12.50 & 69.38 & 3.04 & 7.58 & 4.53 \\
     & L1 & 60.63 & 6.25 & 54.38 & 0.53 & 0.04 & 0.50 & 71.25 & 6.25 & 65.00 & 3.95 & 8.22 & 4.27 \\
     & L2 & 53.75 & 6.25 & 47.50 & 0.46 & 0.03 & 0.42 & \textbf{60.63} & 6.25 & 54.38 & 4.98 & 8.50 & 3.52 \\
    \hline
    InternVLA-N1~\cite{internnav} & L0 & 58.75 & 12.50 & 46.25 & 0.42 & 0.11 & 0.31 & 70.00 & 12.50 & 57.50 & 3.35 & 5.45 & 2.10 \\
     & L1 & 44.38 & 12.50 & 31.88 & 0.34 & 0.11 & 0.23 & 52.50 & 12.50 & 40.00 & 4.53 & 5.99 & 1.46 \\
     & L2 & 33.13 & 0.00 & 33.13 & 0.28 & 0.00 & 0.28 & 43.13 & 0.00 & 43.13 & 5.17 & 7.45 & 2.28 \\
    \hline
    \textbf{HaltNav (Ours)} & L0 & 79.38 & \textbf{50.00} & 29.38 & 0.73 & \textbf{0.44} & 0.29 & 81.88 & \textbf{50.00} & 31.88 & 2.53 & 6.10 & 3.57 \\
     & L1 & 66.25 & \textbf{43.75} & 22.50 & 0.62 &\textbf{0.40}& 0.22 & 71.25 & \textbf{43.75} & 27.50 & 3.52 & 7.01 & 3.49 \\
     & L2 & \textbf{55.63} & \textbf{31.25} & 24.38 & \textbf{0.52} & \textbf{0.29} & 0.23 & 57.50 & \textbf{37.50} & 26.25 & 4.61 & 9.16 & 4.55 \\
    \hline
  \end{tabular}
\end{table*}


\begin{figure*}[t]
  \centering
  \includegraphics[width=0.9\textwidth]{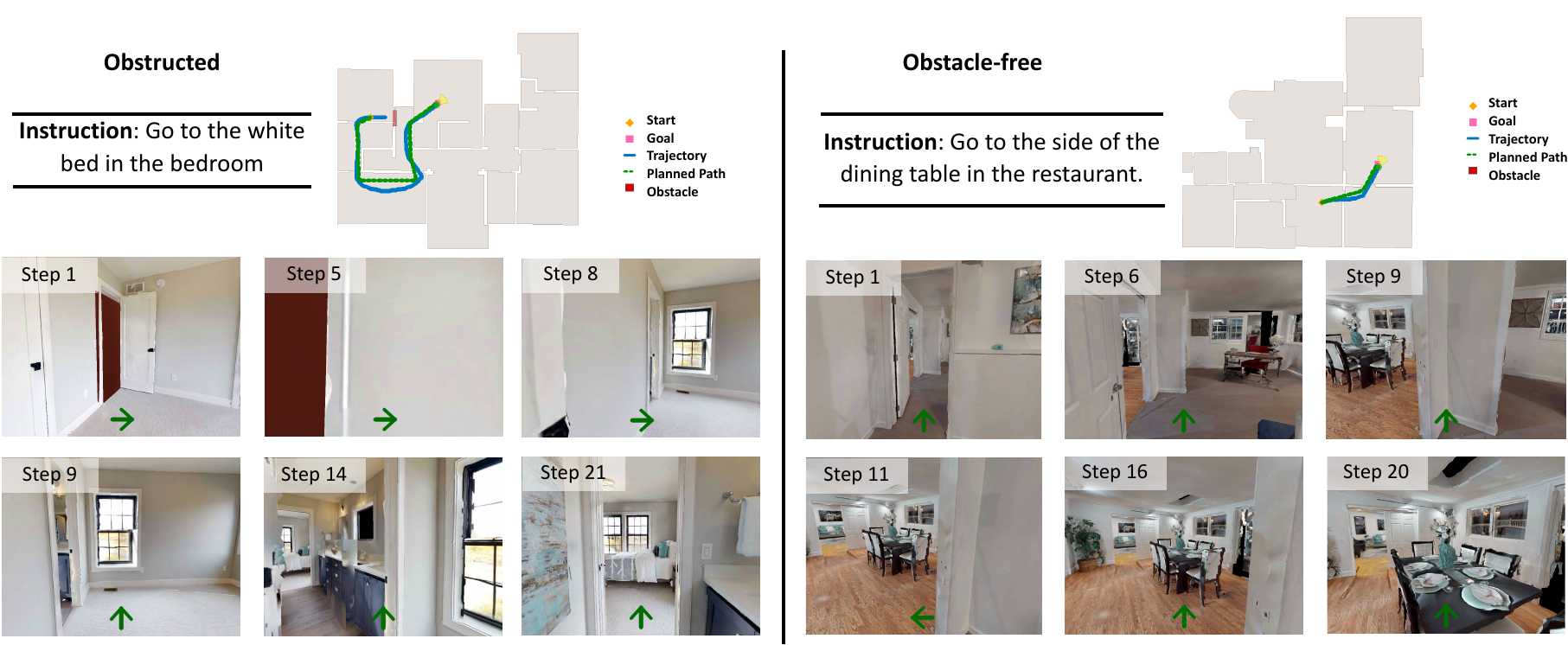}
  \caption{Examples of our proposed methods in HM3D \cite{ramakrishnan2021hm3d} dataset. The left example illustrates the reactive halting and re-planning ability of our pipeline, given the unexpected obstacle observed. The right one shows an ordinary task without injecting any obstacles.}
  \label{fig:example_sim}
\end{figure*}


\section{EXPERIMENTS}

\subsection{Experimental Setup}

To demonstrate the performance of our method, we evaluate the proposed HaltNav framework both in simulators and on a physical robot. Both tests include the standard vision-language navigation tasks and obstacle injection scenarios. 
We also construct a carefully labeled and crafted dataset based on Habitat-Matterport 3D Research Dataset (HM3D) \cite{ramakrishnan2021hm3d}. In each case, the natural language instruction specifying a goal object is provided.


\textbf{Simulation Environment and Dataset.}
The simulation experiments are conducted on Habitat simulator, part of the HM3D toolkits. To evaluate our method, we select 5 diverse indoor scenes from the HM3D. In order to integrate with the osmAG map format, similar to the work in \cite{osmag_llm}, we manually label and create a custom dataset that is compatible with both osmAG map and Habitat environment. It comprises 176 physical navigation tasks, categorized into three difficulty levels: 66 easy, 94 medium, and 16 obstacle tasks. Easy tasks require navigating to another adjacent room, whereas medium tasks are about traversing through multiple rooms. In tasks with obstacles, we add an obstacle to block one of the doors on the shortest path to the goal, which is also the path the language instruction describes. In this case, the robot has to replan a path to the goal. For each individual physical task, there are three different levels of language instructions, effectively tripling the total number of tasks to 528. Level 0 provides the most detailed description including the detailed actions and landmarks. Level 1 primarily describes the landmarks. Level 2 only contains the target object and the target room if the target room has distinctive visual features. 

\textbf{Real-World Platform.} 
\begin{table*}[t]
  \caption{Real-world results on a Fetch robot under standard (B) and obstacle-injected (O) settings. L0 (detailed) and L2 (goal-only) are reported to contrast instruction granularity extremes. For SR/SPL/OS, Drop$=$B$-$O (lower is better). For NE, $\Delta$=O$-$B (lower is better). \textbf{Bold} marks the best SR, SPL and OS under B and O per level.}
  \label{tab:real_world}
  \centering
  \setlength{\tabcolsep}{4.5pt}
  \renewcommand{\arraystretch}{1.05}
  \begin{tabular}{l|c|ccc|ccc|ccc|ccc}
    \hline
    \multirow{2}{*}{Method} & \multirow{2}{*}{Level} & \multicolumn{3}{c|}{SR (\%)} & \multicolumn{3}{c|}{SPL} & \multicolumn{3}{c|}{OS (\%)} & \multicolumn{3}{c}{NE (m)} \\
    \cline{3-14}
     &  & B & O & Drop & B & O & Drop & B & O & Drop & B & O & $\Delta$ \\
    \hline
    StreamVLN~\cite{wei2025streamvln} & L0 & 33.33 & 0.00 & 33.33 & 0.33 & 0.00 & 0.33 & 40.00 & 0.00 & 40.00 & 15.20 & 18.80 & 3.60 \\
     & L2 & 13.33 & 0.00 & 13.33 & 0.13 & 0.00 & 0.13 & 13.33 & 0.00 & 13.33 & 12.05 & 19.58 & 7.53 \\
    \hline
    InternVLA-N1~\cite{internnav} & L0 & 40.00 & 0.00 & 40.00 & 0.39 & 0.00 & 0.39 & 53.33 & 0.00 & 53.33 & 6.86 & 19.94 & 13.08 \\
     & L2 & 0.00 & 0.00 & 0.00 & 0.00 & 0.00 & 0.00 & 0.00 & 0.00 & 0.00 & 27.14 & 25.76 & $-$1.38 \\
    \hline
    \textbf{HaltNav (Ours)} & L0 & \textbf{73.33} & \textbf{56.66} & 16.67 & \textbf{0.72} & \textbf{0.56} & 0.16 & \textbf{86.66} & \textbf{66.66} & 20.00 & 5.31 & 9.25 & 3.94 \\
     & L2 & \textbf{53.33} & \textbf{46.66} & 6.67 & \textbf{0.52} & \textbf{0.46} & 0.06 & \textbf{60.00} & \textbf{46.66} & 13.34 & 11.57 & 14.13 & 2.56 \\
    \hline
  \end{tabular}
\end{table*}
For real-world validation, we deploy our framework on a Fetch robot equipped with an Intel D435 RGB-D camera. The real-world tests are divided into two categories: obstacle-free tasks and tasks with obstacles. In the obstacle tasks, we replicate the simulation setup by physically blocking the doorway along the robot's planned path, forcing it to replan an alternative route to the goal. The experiments are conducted in a university building characterized by long corridors connecting multiple offices and laboratories, presenting significantly greater topological complexity than the simulated home environments.

\textbf{Implementation Details.} The Graph-Grounded Task Dispatcher (GGTD) uses Google Gemini 3 Flash \cite{gemini} to decompose the language instructions into waypoint-level subgoals over the osmAG. The Reactive Visual Halting (RVH) module uses Qwen-2.5-VL-7B \cite{qwen} to monitor the egocentric visual stream and detect navigation anomalies such as blocked passages. The low-level executor is InternVLA-N1~\cite{internnav}, a single-camera VLN policy that converts each subgoal into action commands.

\textbf{Evaluation Metrics.}
We adopt four comprehensive metrics: 
1) \textbf{Success Rate (SR)}: The ratio of episodes where the agent successfully stops within $3.0$ meters of the target.
2) \textbf{Success weighted by Path Length (SPL)}: A standard measure balancing success and trajectory efficiency.
3) \textbf{Oracle Success Rate (OS)}: The percentage of episodes where the agent passes within $3.0$ meters of the target at any point during navigation, regardless of where it ultimately stops.
4) \textbf{Navigation Error (NE)}: The distance between the stop point of the navigation to the ground truth goal position.

\subsection{Simulation Results}
Table~\ref{tab:replanning_bo} compares HaltNav against five baselines across three instruction levels and two scene settings (standard B and obstacle-injected O).

\textbf{Instruction robustness.} As instructions become less specific (L0$\to$L2), all methods degrade, but HaltNav shows the most graceful decline: SR-B drops by 30\% relative (79.38$\to$55.63), compared to 40--44\% for the strongest baselines (OmniNav, InternVLA-N1). The osmAG prior reduces the agent's dependence on detailed route descriptions. We note that OmniNav achieves the highest B-column SR by leveraging panoramic observations from multiple cameras, a setup that substantially increases hardware cost and is infeasible on our single-camera Fetch platform. This constraint is why HaltNav's local VLN executor adopts a single-camera model rather than OmniNav.

\textbf{Replanning robustness.} Under obstacle injection, most baselines fall below 13\% SR at L0-O. HaltNav maintains 50.00\% (L0-O) and 31.25\% (L2-O) with the smallest Drop values across all levels, confirming that RVH effectively detects blockages and triggers viable detours.


\begin{figure}[h]
  \centering
  \includegraphics[width=0.45\textwidth]{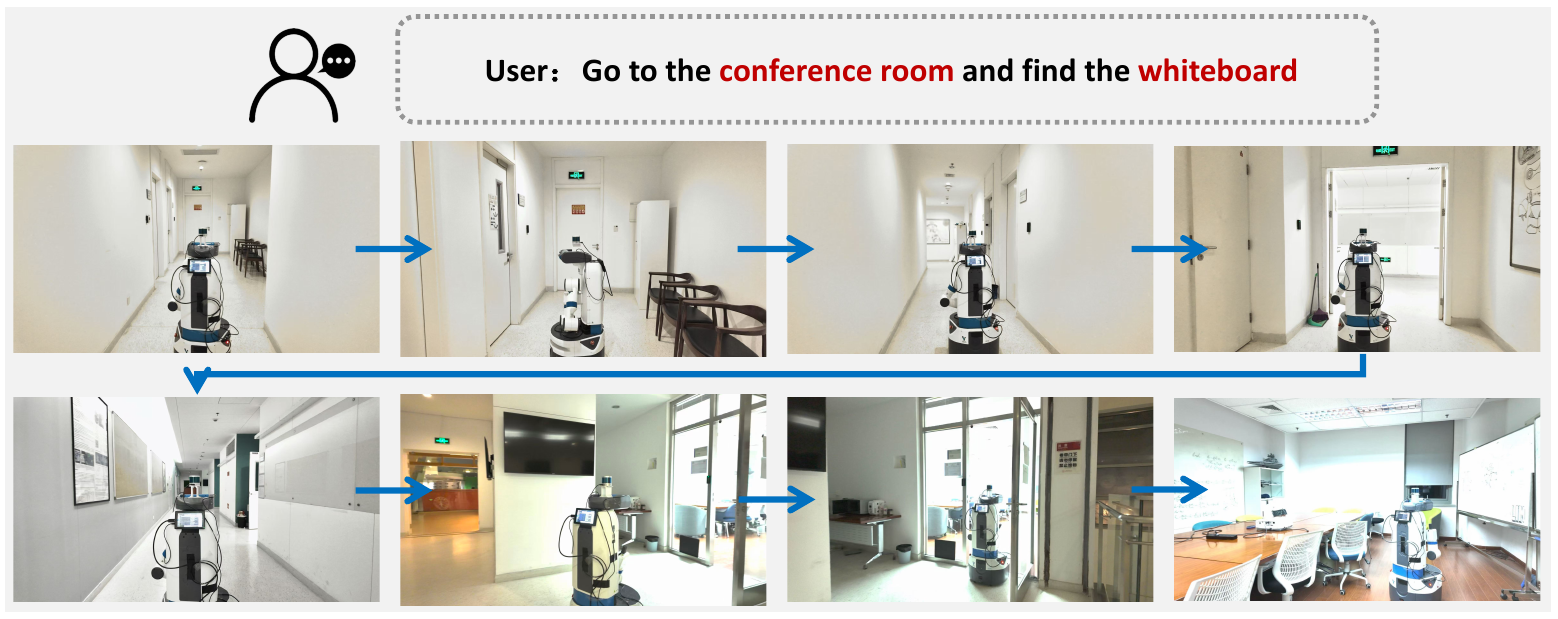}
  \caption{An example of our real world test with the Fetch robot. Given the instruction and osmAG, the robot plans a global path starting in the hallway. However, one of the doors at the end of the hallway on the path is unexpectedly closed. With reactive halting module, the robot identified the situation and replanned via the other doorway on the farther side of the hallway.}
  \label{fig:example_realworld}
\end{figure}

\subsection{Real-World Evaluation Results}
We validate HaltNav on a physical Fetch robot in a university building (Sec.~IV-A). Table~\ref{tab:real_world} reports results for StreamVLN and InternVLA-N1 under L0 (detailed) and L2 (goal-only).

The real-world setting substantially amplifies both failure modes observed in simulation. Under goal-only instructions (L2-B), InternVLA-N1 collapses to 0\% SR and StreamVLN to 13.33\%, far below their simulation counterparts. Under obstacle injection, both baselines drop to 0\% SR across all conditions---even StreamVLN, which retained 37.50\% in simulation. We attribute this to the open, branching topology of real corridors, where a single wrong turn becomes irrecoverable without map guidance. In contrast, HaltNav maintains 56.66\% SR (L0-O) and 46.66\% (L2-O) with SR drops of only 16.67 and 6.67 points, confirming that lightweight topological priors are essential---not merely helpful---for robust real-world VLN when instructions are underspecified and obstacles appear.


%



\section{CONCLUSIONS}
In this work, we presented HaltNav, a hierarchical navigation framework comprising a lightweight osmAG topological prior and an off-the-shelf VLN model. This combination exploits the global planning advantage of the prior map and the local intelligent execution of VLN models. To maximize the utilization of both components at their strengths, we designed an MLLM-based brain that handles task dispatching via the Graph-Grounded Task Dispatcher (GGTD) and detects unexpected anomalies via the Reactive Visual Halting (RVH) mechanism. To fine-tune the brain for robust obstruction understanding, we also developed a dual-engine counterfactual synthesis pipeline that automatically generates scene-based obstacles in navigation images.
Through both simulated and physical robot experiments, we demonstrated that HaltNav outperforms five state-of-the-art baselines. The framework gracefully handles underspecified instructions, and navigates more intelligently when unexpected blockades appear on the planned path. Our real-world experiments further reveal that the advantage of structural priors becomes critical in topologically complex environments, where baselines without map guidance suffer catastrophic failure under goal-only instructions and obstacle injection. To support research on this hybrid system, we have carefully labeled and crafted over 500 high-quality tasks, which will be released upon publication.
        
\FloatBarrier








\bibliographystyle{IEEEtran}
\bibliography{ref}
\end{document}